\documentclass{article}
\usepackage{amsmath,epsfig}
\usepackage[preprint]{spconfa4}
\usepackage{xcolor}
\usepackage{cite}

\usepackage{hyperref}
\usepackage{tabularx}
\usepackage{color}
\usepackage{multirow}
\usepackage{tikz}
\usepackage{amssymb}
\usepackage{longtable, array, booktabs}
\usepackage{makecell}
\usepackage{subfigure}
\usepackage{footnote}
\newcommand\todo[1]{\textcolor{blue}{#1}}

\let\OLDthebibliography\thebibliography
\renewcommand\thebibliography[1]{
  \OLDthebibliography{#1}
  \setlength{\parskip}{0pt}
  \setlength{\itemsep}{0pt plus 0.3ex}
}

\begin{document}\sloppy

\def\x{{\mathbf x}}
\def\L{{\cal L}}

\newcommand\copyrighttext{%
  \footnotesize \textcopyright 2022 IEEE. Personal use of this material is permitted.
  Permission from IEEE must be obtained for all other uses, in any current or future 
  media, including reprinting/republishing this material for advertising or promotional 
  purposes, creating new collective works, for resale or redistribution to servers or 
  lists, or reuse of any copyrighted component of this work in other works. }
  
    \newcommand\mycopyrightnotice{%
\begin{tikzpicture}[remember picture,overlay]
\node[anchor=south,yshift=10pt] at (current page.south) {\fbox{\parbox{\dimexpr\textwidth-\fboxsep-\fboxrule\relax}{\copyrighttext}}};
\end{tikzpicture}%
}

\title{Boosting Video Object segmentation based on scale inconsistency}
%
\name{Hengyi Wang and Changjae Oh}
\address{School of Electronic Engineering and Computer Science, Queen Mary University of London
}

\maketitle

\begin{abstract}
We present a refinement framework to boost the performance of pre-trained semi-supervised video object segmentation (VOS) models. Our work is based on scale inconsistency, which is motivated by the observation that existing VOS models generate inconsistent predictions from input frames with different sizes. We use the scale inconsistency as a clue to devise a pixel-level attention module that aggregates the advantages of the predictions from different-size inputs. The scale inconsistency is also used to regularize the training based on a pixel-level variance measured by an uncertainty estimation. We further present a self-supervised online adaptation, tailored for test-time optimization, that bootstraps the predictions without ground-truth masks based on the scale inconsistency. Experiments on DAVIS 16 and DAVIS 17 datasets show that our framework can be generically applied to various VOS models and improve their performance.
\end{abstract}
\mycopyrightnotice
\section{Introduction}
\label{sec:intro}

Video object segmentation (VOS) aims to divide target objects from other instances in a video sequence. In this work, we focus on a semi-supervised setting, where the ground-truth mask of the first frame is given. Semi-supervised VOS~\cite{perazzi2016benchmark,pont20172017} is challenging as the model needs to address appearance changes, similar instances, occlusions, and scale variations based on the mask of the first frame. 

Existing deep learning-based methods to address the aforementioned problems can be categorized into three approaches. \textit{Online learning-based} methods fine-tune a model on the ground-truth mask of the first frame at test time~\cite{caelles2017one,voigtlaender2017online,maninis2018video,meinhardt2020make}. \textit{Propagation-based} methods use predicted masks from the past frames to guide the current prediction~\cite{perazzi2017learning,jampani2017video,xiao2018monet,oh2018fast,khoreva2019lucid}, and 
\textit{Matching-based} methods perform feature matching in embedding space to segment the target object~\cite{wang2019ranet,oh2019video,yang2020collaborative,seong2021hierarchical,cheng2021stcn}. 
These methods mainly focus on designing networks to improve segmentation accuracy.

\begin{figure}
\centering
\begin{tabular}{c@{\hspace{2pt}}c@{\hspace{2pt}}c@{\hspace{2pt}}c}
\includegraphics[width=0.24\columnwidth]{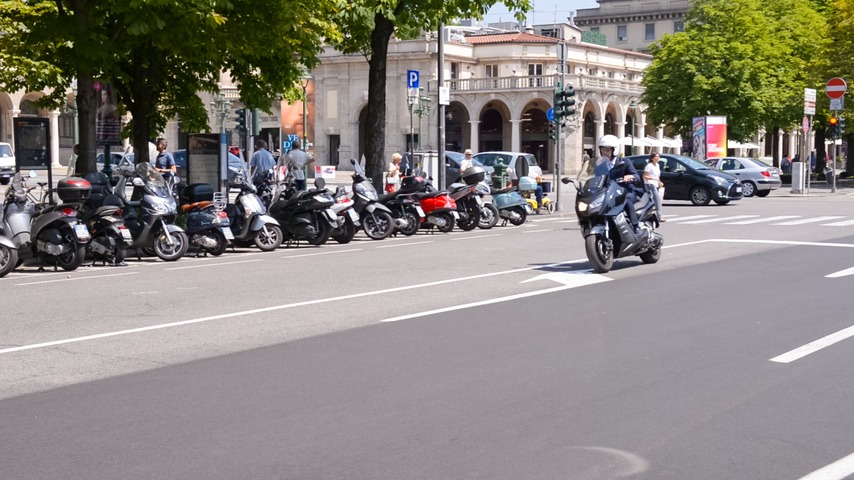}
&\includegraphics[width=0.24\columnwidth]{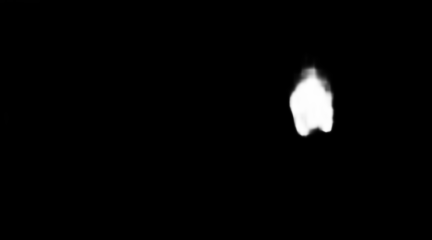}
&\includegraphics[width=0.24\columnwidth]{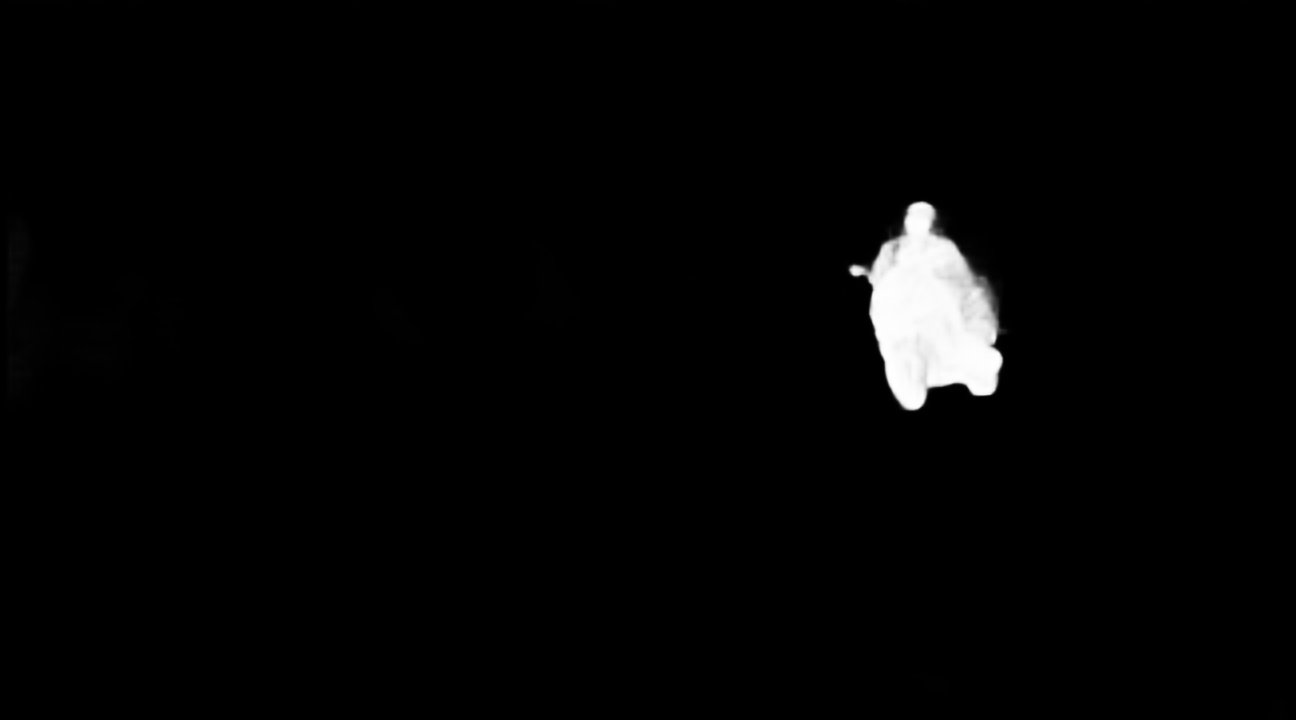}
&\includegraphics[width=0.24\columnwidth]{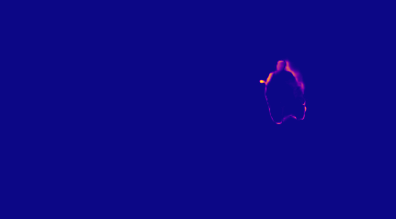}\\

\includegraphics[width=0.24\columnwidth]{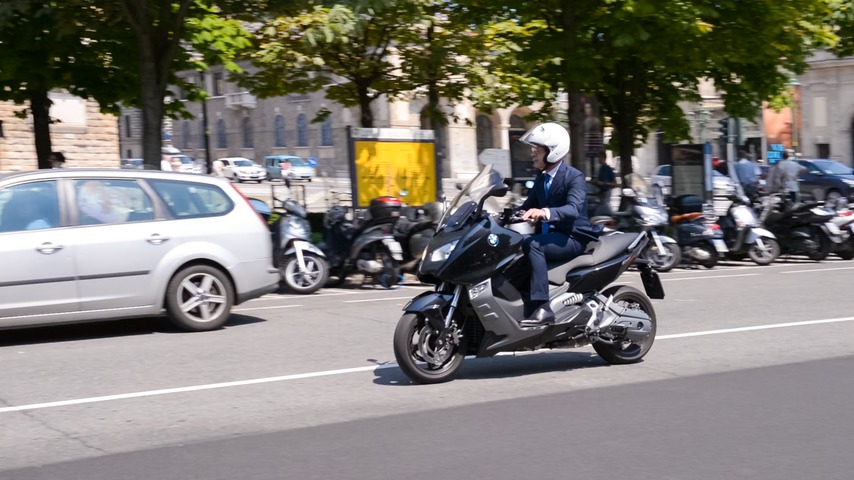}
&\includegraphics[width=0.24\columnwidth]{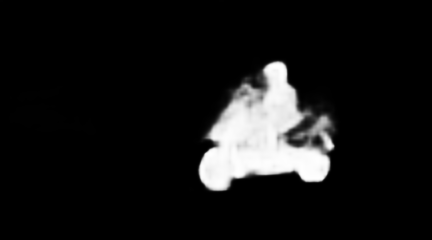}
&\includegraphics[width=0.24\columnwidth]{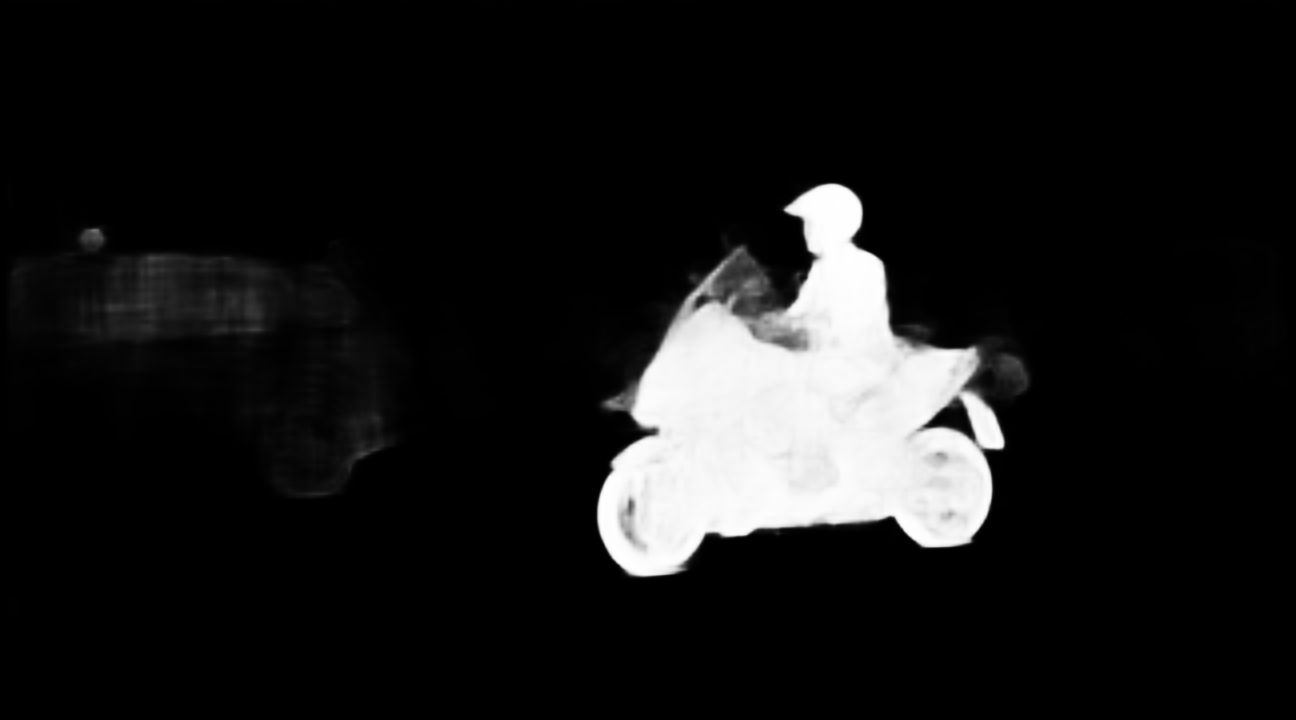}
&\includegraphics[width=0.24\columnwidth]{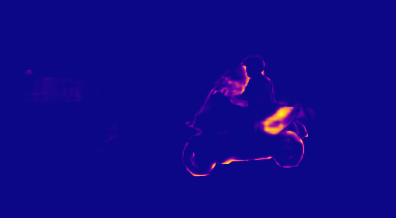}\\

\includegraphics[width=0.24\columnwidth]{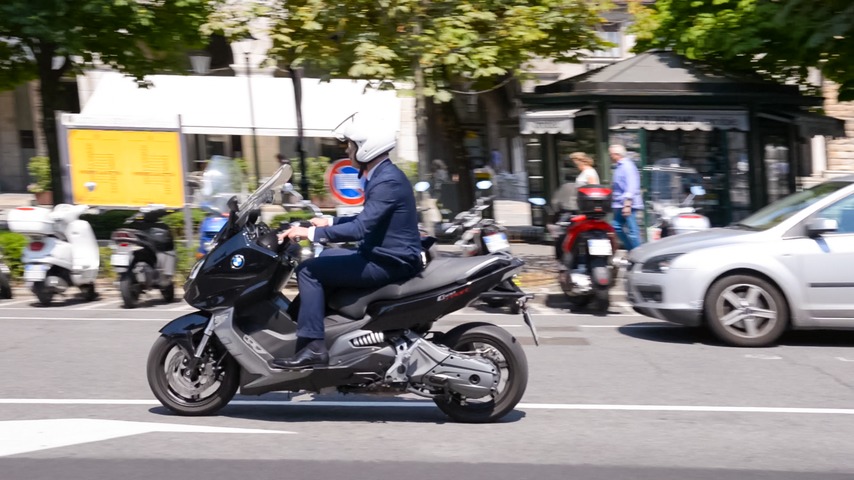}
&\includegraphics[width=0.24\columnwidth]{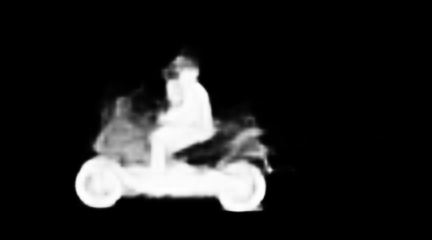}
&\includegraphics[width=0.24\columnwidth]{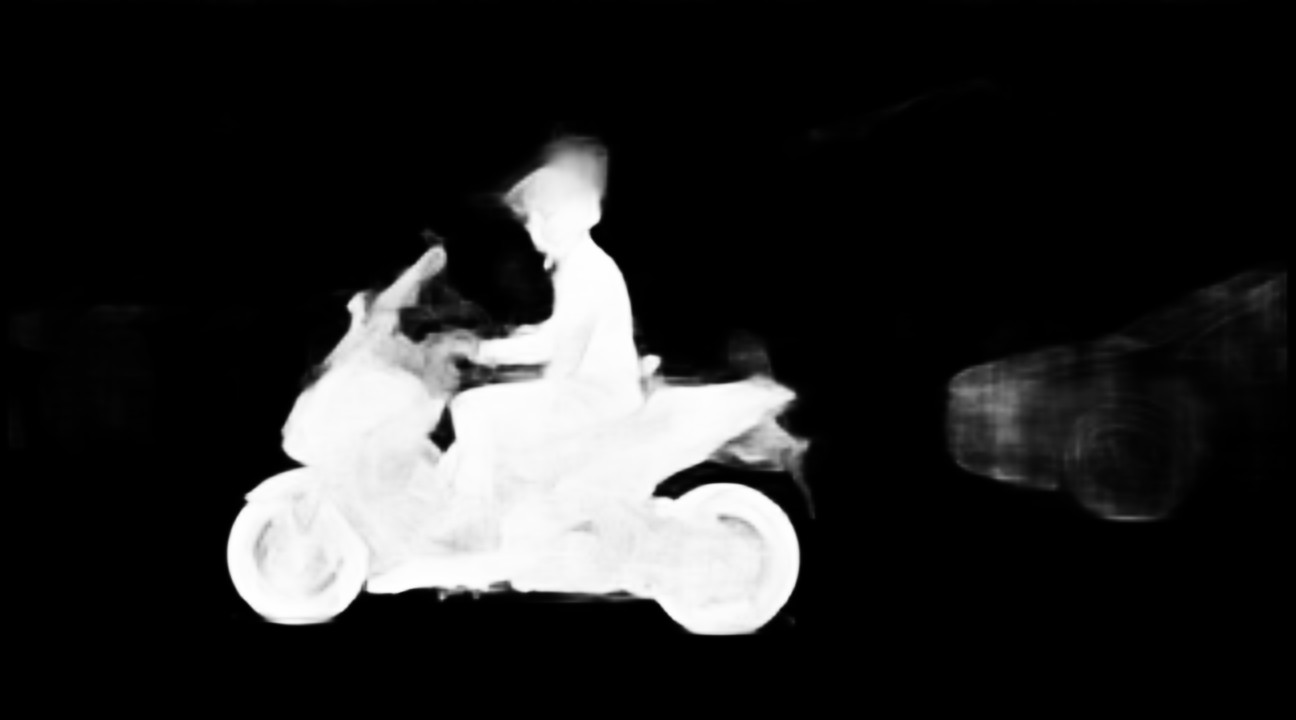}
&\includegraphics[width=0.24\columnwidth]{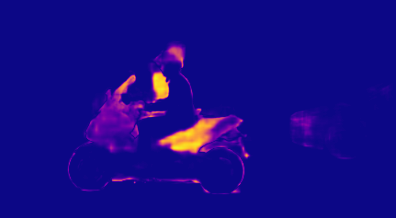}\\

 \footnotesize{(a)}
 &\footnotesize{(b)}
 &\footnotesize{(c)}
 &\footnotesize{(d)} \\
\end{tabular}
\vspace{-9pt}
\caption{ Scale inconsistency in video object segmentation. (a) Input frames, predictions from (b) small-size ($\times 0.5$) and (c) original-size ($\times 1.0$) inputs, and (c) their variance maps showing scale inconsistency. We use this scale inconsistency to improve the performance of pre-trained VOS models.} 
\label{fig:introduction}
\vspace{-9pt}
\end{figure}

We observe that existing VOS models commonly generate inconsistent predictions when the same frames with different sizes are used as input. As shown in Fig.~\ref{fig:introduction}, the predictions from the different-size inputs (Figs.~\ref{fig:introduction}(b) and~\ref{fig:introduction}(c)) show the inconsistent results. Some methods~\cite{oh2018fast,yang2019anchor} address the scale-inconsistency problem by averaging the predictions of the inputs with multiple sizes. However, the simple averaging cannot fully address the scale-inconsistency as the magnitude of the scale inconsistency varies from pixel to pixel and the amount of the inconsistencies are different for each frame in a video sequence (Fig.~\ref{fig:introduction}(d)).

\begin{figure*}[t]
    \centering
    \begin{tabular}{cc}
    \includegraphics[height=0.15\textheight]{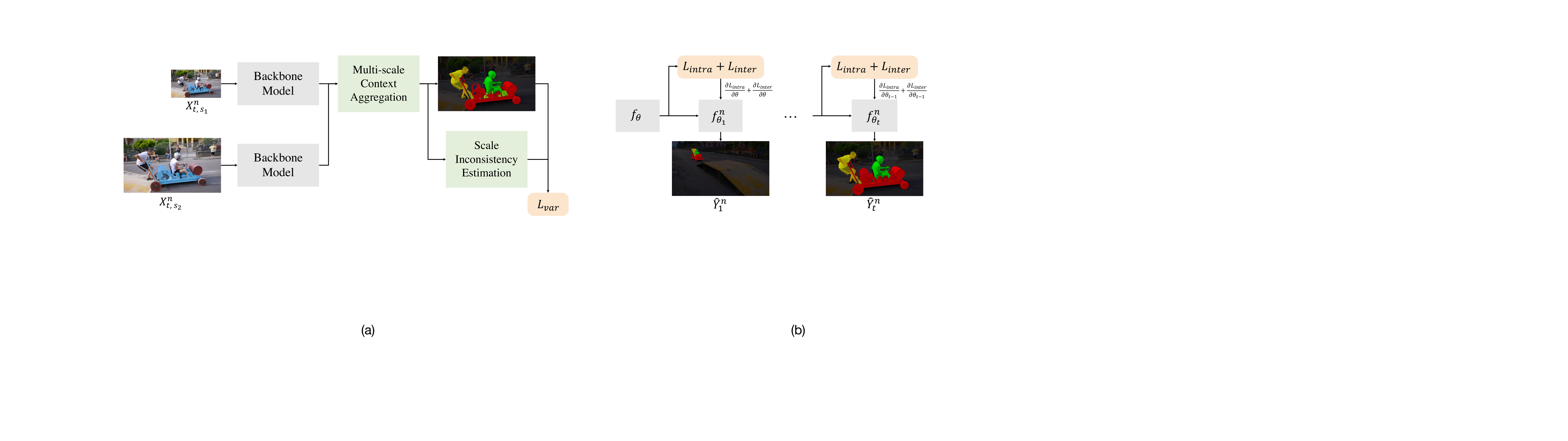}
    &\includegraphics[height=0.15\textheight]{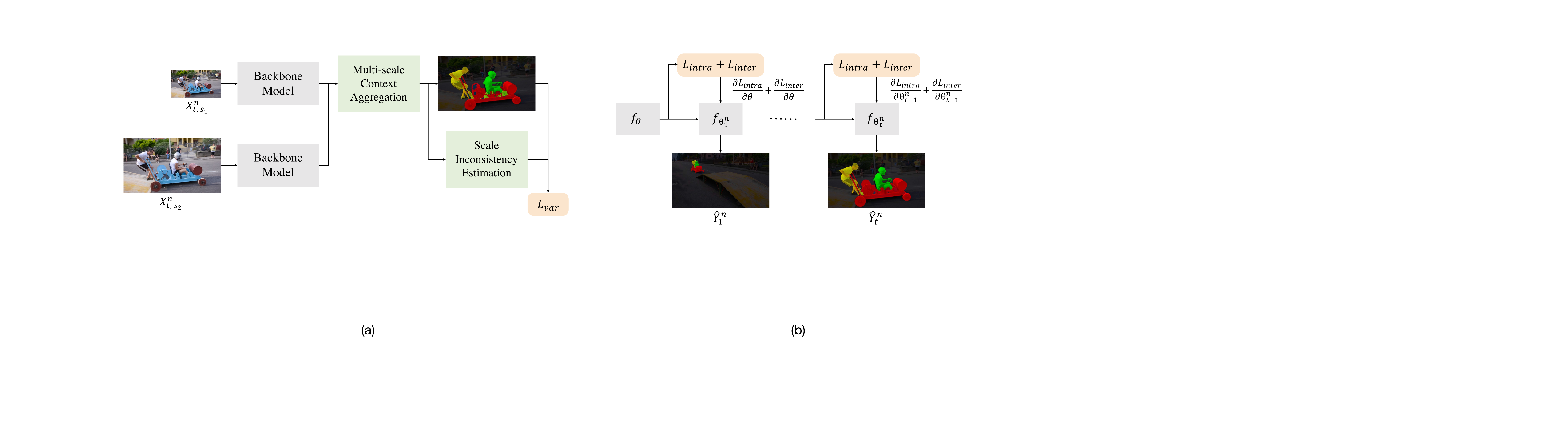}\\
    \footnotesize{(a) Offline training}
    &\footnotesize{(b) Online adaptation}
    \end{tabular}
    \caption{Overview of the proposed refinement framework. During the (a) offline training, we employ a multi-scale context aggregation module with a scale inconsistency estimation to optimize the model to capture the predictions from the multi-scale inputs. The training is optimized by the variance-based segmentation loss $L_{var}$ that can reduce the scale inconsistency. (b) At test time, we perform the online adaptation on model $f_{\theta}$ obtained from the offline training. For the $t$-th frame in the $n$-th video sequence, the model is updated to $f_{\theta_t^n}$, based on the loss functions for inter-frame ($L_{inter}$) and intra-frame ($L_{intra}$) adaptation, and outputs the final prediction $\hat{Y}_t^n$ at time $t$. A backbone model can be replaced with various pre-trained VOS models.}
    \label{fig:Overview}
\end{figure*}

In this paper, we propose a model-agnostic refinement framework that improves existing VOS models by addressing the aforementioned scale-inconsistency problem.\footnote{Project page: \todo{\url{https://hengyiwang.github.io/projects/icme22.html}}} We devise a multi-scale context aggregation module that combines the predictions from different-size inputs using learnable pixel-level attention. We train this module by constructing a pixel-level variance map based on the scale inconsistency from the multi-scale predictions. The pixel-level variance map is used for regularizing a segmentation loss, i.e. a pixel with larger-scale inconsistency is more penalized during the training. To prevent the false-positive error accumulation at the test time, we further present a self-supervised online adaptation that optimizes the model parameters based on scale inconsistency. Specifically, the proposed online adaptation consists of an intra-frame adaptation, which performs pseudo-label learning based on the pixel-level variance map, and an inter-frame adaptation, which enforces the consistency between the adjacent frames using color and distance cues. The experiments on the DAVIS16~\cite{perazzi2016benchmark} and DAVIS17~\cite{pont20172017} datasets show that the proposed method improves the $\mathcal{J\&F}$ measure of three representative VOS models, OSVOS~\cite{caelles2017one}, RGMP~\cite{oh2018fast}, STM~\cite{oh2019video}, by $7.0\%$, $1.1\%$, $2.1\%$ in DAVIS16 and $12.3\%$, $0.7\%$, $1.8\%$ in DAVIS17, respectively.
\section{Problem Statement}
Let us denote an $n$-th video sequence with multiple frames, where the $t$-th frame is $X_{t}^{n}$. 
Semi-supervised VOS aims to predict the masks $\hat{Y}_{t}^{n}$ corresponding to $X_{t}^{n}$ with the annotated object mask from the first frame, ${Y}_{0}^{n}$. In this section, we describe the deep learning pipeline for VOS models with three stages: offline training, online learning, and online adaptation.
\textit{Offline training} aims at training a VOS model $f_{\theta}$, with learnable parameters $\theta$, on a training dataset to learn how to segment a target object from the background. The objective function can be formulated as:
\begin{equation}
	\theta=\mathop{\arg\min}_{\theta}{\sum\limits_{n}\sum\limits_{t}{L}_{seg}\left(f_{\theta}\left(X_{t}^{n}\right), Y_{t}^{n}\right)},
\label{offline}
\end{equation}
\textit{Online learning} fine-tunes $f_{\theta}$ on the annotated object mask $Y_{0}^{n}$ of the first frame to learn the specific target semantics and infer the masks of the rest frames with the learned parameters. The parameters in the VOS model are thus obtained for each video sequence to segment the target object. Hence, the model parameters, $\theta^{n}$, for the $n$-th video can be obtained as follows:
\begin{equation}
	\theta^{n}=\mathop{\arg\min}_{\theta^{n}}{{L}_{seg}\left(f_{\theta^{n}}\left(X_{0}^{n}\right), Y_{0}^{n}\right)}.
\label{online}
\end{equation}
\textit{Online adaptation} updates $f_{\theta}$ during the test time by learning from the frames without ground-truth annotations. At time $t$, the model parameter, $\theta_{t}$, for this frame is required to accurately predict the mask. The poor adaptability issue in offline training and online learning can be addressed by online adaptation. For the $n$-th video sequence, the model parameters $\theta_{t}^{n}$ at time $t$ can be obtained as follows:
\begin{equation}
	\theta_{t}^{n}=\mathop{\arg\min}_{\theta_{t-1}^{n}}{{L}_{seg}\left(f_{\theta_{t-1}^{n}}\left(X_{t}^{n}\right), \hat{Y}_{t}^{n}\right)}.
\label{OnlineAda}
\end{equation}
In our approach, we mainly focus on the phases in Eq.~\ref{offline} and Eq.~\ref{OnlineAda} to improve existing semi-supervised VOS models.
\section{Proposed Method}
Fig.~\ref{fig:Overview} shows an overview of the proposed refinement framework. We aim to improve pre-trained VOS models (backbone model) using the multi-scale context aggregation module with the scale inconsistency estimation (Fig.~\ref{fig:Overview} (a)). At test-time, we further perform the self-supervised online adaptation that updates the model parameters to reduce the accumulation of the errors over the frames, caused by the increase of the scale inconsistency (Fig.~\ref{fig:Overview} (b)).
\subsection{Multi-scale Context Aggregation Module}
\begin{figure}
\centering
\begin{tabular}{c@{\hspace{2pt}}c@{\hspace{2pt}}c@{\hspace{2pt}}c}
\includegraphics[width=0.31\columnwidth]{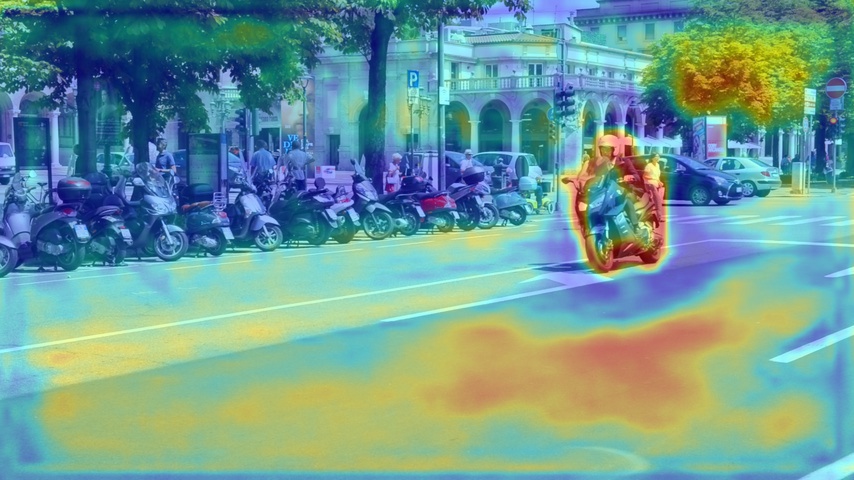}
\includegraphics[width=0.31\columnwidth]{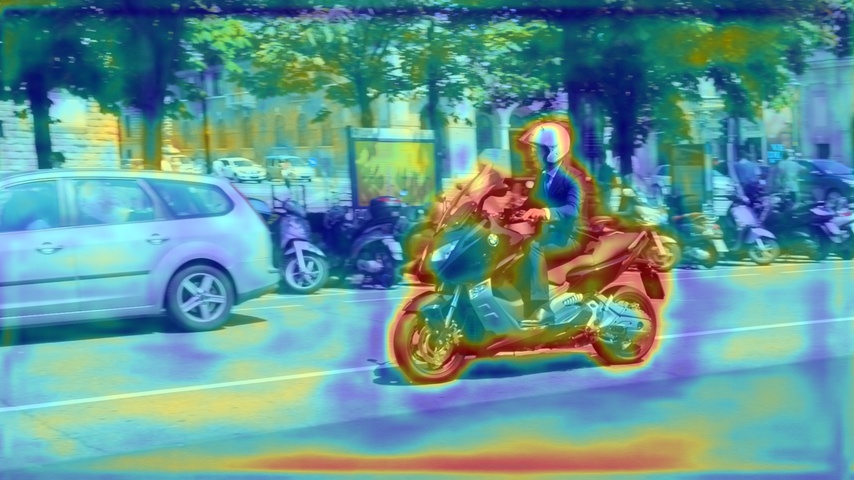}
&\includegraphics[width=0.31\columnwidth]{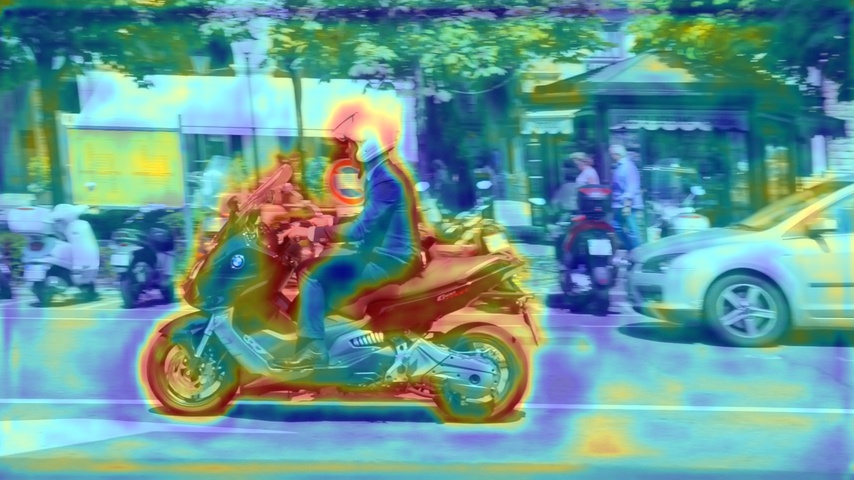}
\end{tabular}
\caption{Visual examples of the pixel-level attention map $A_{t}^{n}$. $A_{t}^{n}$ is color-coded from red (large) to blue (small), where the red pixels have larger fusion weight on the predictions from the larger-size input $X_{t,s_2}^n$.} 
\label{fig:attention}
\end{figure}

We present a multi-scale context aggregation module to learn pixel-wise attention that provides the fusion weight between multi-scale predictions. We use different-size inputs, $X_t^n$ with scale $s_1$ and $s_2$, for a backbone VOS model. The feature maps of different-size inputs before the last convolution layer are extracted, resized, and concatenated to generate a pixel-wise attention map $A_t^n$ (see Fig.~\ref{fig:attention}) by our attention module, which only consists of three convolution layers. $A_{t}^{n}$ is then used as a fusion weight to combine the intermediate predictions $M_{t, s_{1}}^{n}$ and $M_{t, s_{2}}^{n}$ as follows:
\begin{equation}
\hat{Y}_{t}^{n} = \mathcal{U}\left(\mathcal{U}(A_{t}^{n}, s_1) \circ M_{t, s_{1}}^{n}, s_2\right) + \left(\mathbf{1} - \mathcal{U}(A_{t}^{n}, s_2)\right) \circ M_{t, s_{2}}^{n},
\label{Eq.MsAttn}
\end{equation}
\noindent
where $\circ$ is element-wise multiplication and $\mathcal{U}(A_{t}^{n}, s_2)$ denotes that the attention map $A_{t}^{n}$ is upsampled to the scale $s_2$. Our multi-scale context aggregation not only improves the performance of the backbone models on occlusion or fine structures but is robust to maintain the object from other similar instances or appearance changes. 

A straightforward approach to address the scale-inconsistency problem is averaging the multi-scale predictions~\cite{oh2018fast,yang2019anchor}. However, as demonstrated by the works in image segmentation~\cite{chen2016attention,tao2020hierarchical}, simply averaging the predictions cannot effectively handle the scale inconsistency as the magnitude of inconsistency differs from pixel to pixel. Unlike existing methods, our method with the learnable attention map provides different fusion weights to each pixel, which can effectively combine the multi-scale context. 

\subsection{Scale Inconsistency-based Variance Regularization}
Our model incorporates a different amount of context from multi-scale inputs. These multi-scale predictions thus can include the region with large inconsistencies, where the uncertainty of the predictions is high, e.g. pixels with large inconsistencies are prone to cause false predictions. We address this issue by estimating the pixel-wise scale inconsistency, i.e. uncertainty, $V_{t}^{n}(\mathbf{p})$ at point $\mathbf{p}$, using KL divergence:  
\begin{equation}
V_{t}^{n}(\mathbf{p})=\mathcal{E}\left[M_{t, s_{1}}^{n}(\mathbf{p})\cdot \log \left(\frac{\mathcal{U}(M_{t, s_{1}}^{n}, s_2)(\mathbf{p})}{M_{t, s_{2}}^{n}(\mathbf{p})}\right)\right].
\label{KLDiv}
\end{equation}
\noindent
The variance map $V_{t}^{n}$ represents the uncertainty of each pixel. To regularize the segmentation loss, $L_{seg}$, we encourage the model to focus more on the regions with large variance as well as minimizing the scale inconsistency, as follows:
\begin{equation}
L_{var}= \sum_{\mathbf{p}} e^{\beta V_{t}^{n}(\mathbf{p})} L_{seg}(\mathbf{p}),
\label{newLoss}
\end{equation}
\noindent
where $\beta$ controls the effect of $V_{t}^{n}$. By setting $\beta>0$, in offline training, the model can focus more on pixels that are difficult to predict the result. 
\subsection{Scale inconsistency-based Online adaptation}
Multi-scale prediction commonly introduces some false-positive errors, which can be critical to semi-supervised VOS as the errors from the past frames are likely to be accumulated. To address this problem, we propose an online adaptation method that aims at suppressing the error accumulation at test time. At test time, our online adaption updates the model parameters by considering the intra-frame and inter-frame adaptation. Both of these adaptations are performed in a self-supervised manner that does not require the ground-truth mask for updating the model parameters.

\noindent
\textbf{Intra-frame adaptation.} Given the current noisy prediction $\hat{Y}_t^n$, we can learn from pixels with high confidence based on the scale inconsistency variance map ${V}_t^n$, as follows: 
\begin{equation}
L_{intra}= \sum_{\mathbf{p}} e^{-V_{t}^{n}(\mathbf{p})} L_{seg}(\mathbf{p}).
\label{intraLoss}
\end{equation}
Smaller weight is assigned to the pixels with higher variance, as these pixels can generate inaccurate predictions. Note that the $V_t^n$ in $L_{intra}$ is not trainable, i.e. we use the fixed parameters, learned from Eq.~\ref{newLoss}, only to guide the training. Our intra-frame adaptation is inspired by~\cite{zheng2021rectifying} which introduces an auxiliary classifier for automatic pseudo-label learning. Unlike~\cite{zheng2021rectifying} that requires extra parameters for the adaptation, the proposed approach utilizes the scale inconsistency of VOS models to naturally provide the training weight for the intra-frame adaptation.
\begin{figure}[t]
    \centering
    \includegraphics[width=\columnwidth]{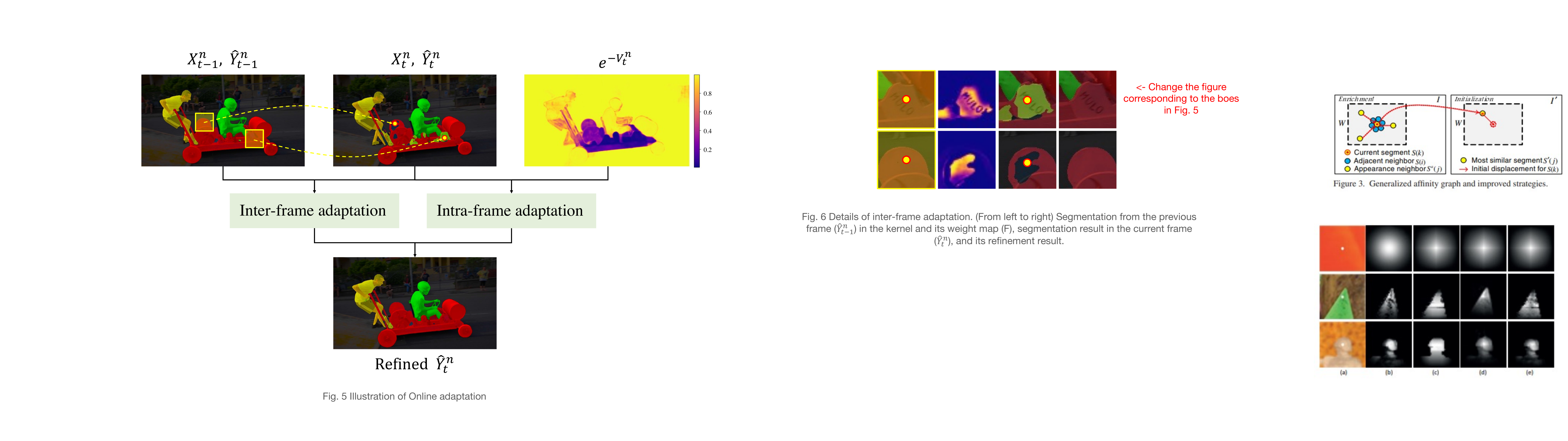} 
    \caption{Illustration of the proposed online adaptation.}
    \label{fig:onlineAda}
\end{figure}

\begin{figure}[t]
    \centering
    \includegraphics[width=0.7\columnwidth]{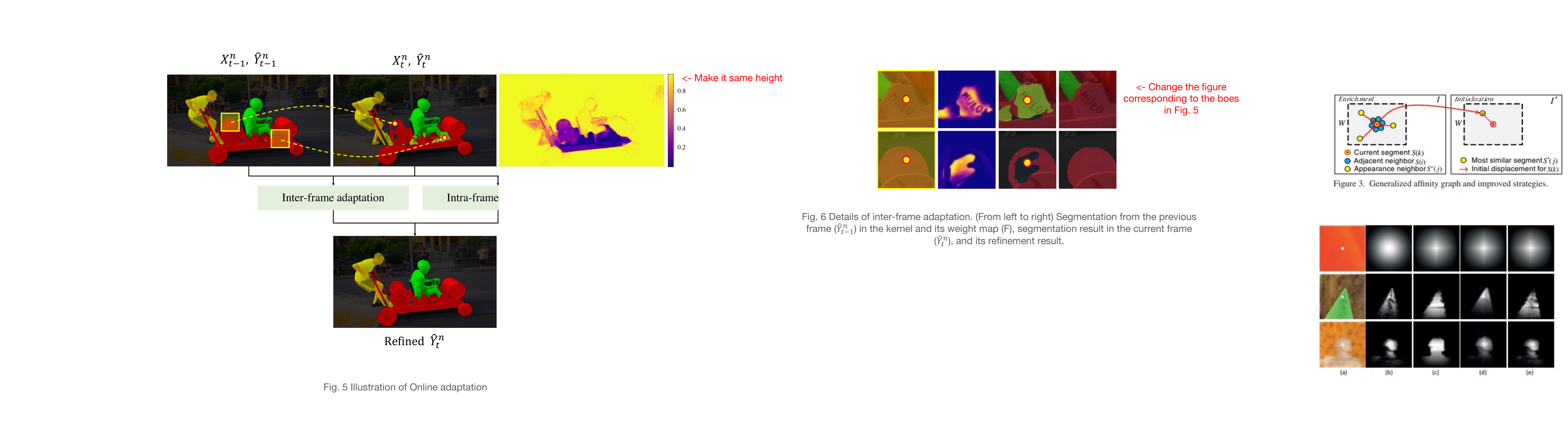} 
    \caption{Details of the inter-frame adaptation. (From left to right) The segmentation from the previous frame, $\hat{Y}_{t-1}^{n}$, the visualization of $L_{inter}$, the result in the current frame $\hat{Y}_{t}^{n}$, and the refinement result. Images are cropped from Fig.~\ref{fig:onlineAda}.}
    \label{fig:inter}
\end{figure}

\noindent
\textbf{Inter-frame adaptation.} The proposed intra-frame adaptation can well exploit the information within a frame and keep adapting the model on each frame independently. To consider the temporal information between the frames, we further present an inter-frame adaptation that encourages consistent predictions between adjacent frames. For the point $\mathbf{p}$ in frame $t$, we set a $k\times k$ kernel $K_{\mathbf{p}}$ in frame $t-1$ in which its center is located to the same position as $\mathbf{p}$, assuming that the displacement between adjacent frames is small. The points $\mathbf{q}$ in $K_{\mathbf{p}}$ are used to determine the label of $\mathbf{p}$ in frame $t$ by aggregating the similarity between $\mathbf{p}$ and $\mathbf{q}$ as follows:
\begin{equation}
L_{inter}=\sum_{\mathbf{p}} \sum_{\mathbf{q} \in K_{\mathbf{p}}} F(\mathbf{p}, \mathbf{q}) D(\mathbf{p}, \mathbf{q}),
\end{equation} 
\noindent
where $D(\mathbf{p}, \mathbf{q})=\left|\hat{Y}_{t}^{n}(\mathbf{p})-\hat{Y}_{t-1}^{n}(\mathbf{q})\right|$ measures the absolute difference between the label of point $\mathbf{p}$ at frame $t$ and $\mathbf{q}$ at frame $t-1$. Namely, $L_{inter}$ measures the difference between the point $\mathbf{p}$ in the current frame and its neighbor pixels in the previous frame by considering the spatial and intensity distance with $F(\mathbf{p}, \mathbf{q})$:
\begin{equation}
F(\mathbf{p}, \mathbf{q})=\frac{1}{w} \exp \left(-\frac{\|\mathbf{p}-\mathbf{q}\|^{2}}{2 \sigma_{P}^{2}}-\frac{\|I_{t}(\mathbf{p})-I_{t-1}(\mathbf{q})\|^{2}}{2 \sigma_{I}^{2}}\right),
\end{equation}
\noindent
where $w$ is a normalization coefficient, and the parameters $\sigma_{P}^{2}$ and $\sigma_{I}^{2}$ are considered as the spatial and intensity variance. $I_{t}(\cdot)$ and $I_{t-1}(\cdot)$ are the RGB value of the frame at $t$ and $t-1$, respectively.
As shown in Fig.~\ref{fig:inter}, our inter-frame adaptation measures the weight between $\mathbf{p}$ and all points in the kernel $K_{\mathbf{p}}$ and aggregates the prediction information.

The final objective function of the online adaptation, $L_{online}$, is the combination of the inter-frame adaptation and the intra-frame adaptation as follows:
\begin{equation}
L_{online} = L_{intra} + L_{inter}.
\end{equation}
\noindent
The intra-frame adaptation allows the model to adapt to the current frame while the inter-frame adaptation enforces the temporal consistency, which provides extra supervision for the pixels with high variance. At test time, these two adaptations are jointly used to update the model parameters without ground-truth masks. 
\section{Validation}
We validate our method with three pre-trained backbone models and evaluate the results on the DAVIS 16 \cite{perazzi2016benchmark} and DAVIS 17 \cite{pont20172017} datasets. We also present the ablation analysis to verify the effectiveness of each component in our method.
\subsection{Setup}
\noindent
\textbf{Baselines.} We validate our method using three pre-trained VOS models as backbone, OSVOS~\cite{caelles2017one}, RGMP~\cite{oh2018fast} and STM~\cite{oh2019video}. Each model is a representative in online learning-based, propagation-based, and matching-based methods.

\noindent
\textbf{Datasets.} We adopt DAVIS 16~\cite{perazzi2016benchmark} and DAVIS 17~\cite{pont20172017} dataset to evaluate the proposed method. DAVIS 16 contains a total of 50 video sequences which are divided into 30 training sequences and 20 validation sequences with foreground and background annotations. DAVIS 17 consists of 150 videos in total with instance-level annotations. The dataset is split into 60 training sequences, 30 validation sequences and 30 test sequences. These two datasets are used to evaluate the single-object and multi-object VOS, respectively.

\noindent
\textbf{Evaluation metrics.} We use Jaccard index ($\mathcal{J}$) and F-measure ($\mathcal{F}$) to measure the region similarity and contour accuracy~\cite{pont20172017}. $\mathcal{J}$-Decay and $\mathcal{F}$-Decay denote the performance decay of $\mathcal{J}$ and $\mathcal{F}$ over time. The final $\mathcal{J\&F}$ score is obtained by averaging the value of $\mathcal{J}$ and $\mathcal{F}$.

\noindent
\textbf{Implementation details.} To make a fair comparison with existing methods, we only used DAVIS datasets for offline training. For RGMP and STM, we used their publicly available pre-trained parameters and leverage the same training strategy as their original implementations. For OSVOS, we modified its structure to DeepLabv3+~\cite{chen2018encoder} and trained the model from scratch on the DAVIS dataset. During the online adaptation, we update the model parameters for each frame sequentially.

\begin{table}[t]
\centering
\normalsize
\resizebox{0.99\columnwidth}{!}{
\setlength{\tabcolsep}{2pt}
\begin{tabular}{r c lll lll }

\specialrule{1.2pt}{0.2pt}{1pt}
\multicolumn{2}{c}{Methods}       & \multicolumn{3}{c}{DAVIS 2016 (val)}    & \multicolumn{3}{c}{DAVIS 2017 (val)}    \\ 
\cmidrule(lr){1-2}
\cmidrule(lr){3-5}
\cmidrule(lr){6-8}
Name &     O/P/M               & $\mathcal{J} \uparrow$             & $\mathcal{F} \uparrow$             & $\mathcal{J}$\&$\mathcal{F} \uparrow$           & $\mathcal{J} \uparrow$             & $\mathcal{F} \uparrow$             & $\mathcal{J}$\&$\mathcal{F} \uparrow$           \\ 
\cmidrule(lr){1-8}
OnAVOS~\cite{voigtlaender2017online}        & O &86.1          & 84.9          & 85.5           & 61.6          & 69.1          & 65.4         \\
OSVOS-S~\cite{maninis2018video}       & O &85.6          & 87.5          & 86.6          & 64.7          & 71.3          & 68.0             \\
e-OSVOS~\cite{meinhardt2020make}       & O &86.6 & 87.0            & 86.8           & 74.4 & 80.0            & 77.2           \\ 
MaskTrack~\cite{perazzi2017learning}      & P &79.7          & 75.4          & 77.6          & ——            & ——            & ——             \\
Lucid~\cite{khoreva2019lucid}       & P & 83.9          & 82.0            & 82.9          & ——            & ——            & ——             \\
MHP-VOS~\cite{xu2019mhp}      & P & 87.6 & 89.5 & 88.6 & 73.4 & 78.9 & 76.2 \\ 
CFBI+~\cite{yang2020collaborative}         & M & 88.7          & 91.1          & 89.9           & 80.1          & 85.7          & 82.9           \\
HMMN~\cite{seong2021hierarchical}        & M & 89.6          & 92.0         & 90.8          & 81.9          & 87.5          & 84.7         \\
STCN~\cite{cheng2021stcn}        & M & 90.4          & \textbf{93.0}          & \textbf{91.7}         & \textbf{82.0}          & \textbf{88.6}         & \textbf{85.3}        \\
\cmidrule(lr){1-8}
OSVOS~\cite{caelles2017one}         & O & 79.8          & 80.6          & 80.2           & 56.6          & 63.9          & 60.3          \\
RGMP~\cite{oh2018fast}         & P & $81.5$          & $82.0$            & $81.7$          & 64.8          & 68.6          & 66.7           \\
STM~\cite{oh2019video}          & M & 88.7          & 90.1          & 89.4           & 79.2          & 84.3          & 81.8          \\ 
\cmidrule(lr){1-8}
OSVOS + \textbf{Ours}            & O & $86.4_{{\todo{+6.6}}}$          & $87.9_{{\todo{+7.3}}}$ & $87.2_{{\todo{+7.0}}}$ & $69.7_{{\todo{+13.1}}}$          & $75.5_{{\todo{+11.6}}}$          & $72.6_{{\todo{+12.3}}}$           \\
RGMP + \textbf{Ours}     & P & $83.1_{\todo{+1.6}}$          & $82.4_{\todo{+0.4}}$          & $82.8_{\todo{+1.1}}$          & $65.0_{{+0.2}}$          & $69.7_{\todo{+1.0}}$          & $67.4_{\todo{+0.7}}$          \\
STM + \textbf{Ours}      & M & $\textbf{91.1}_{\todo{+2.4}}$ & $91.9_{\todo{+1.8}}$ & $91.5_{\todo{+2.1}}$  & $81.3_{\todo{+2.1}}$ & $85.9_{\todo{+1.6}}$          & $83.6_{\todo{+1.8}}$  \\ 
\specialrule{1.2pt}{0.2pt}{1pt}
\end{tabular}}
\caption{Evaluation on the DAVIS 16 and DAVIS 17 validation (val) datasets. KEY -- O: Online learning-based, P: Propagation-based, M: Matching-based, $\mathcal{J}$: Jaccard index, $\mathcal{F}$: F-measure, $\uparrow$: the higher, the better.} 
\label{tab:quant}
\end{table}
\begin{figure}[t]
\centering
\begin{tabular}{c@{\hspace{2pt}}r@{\hspace{2pt}}r@{\hspace{2pt}}r@{\hspace{2pt}}r}   
\raisebox{0.2\height}{\rotatebox{90}{\scriptsize OSVOS}}
& \includegraphics[width=0.23\linewidth]{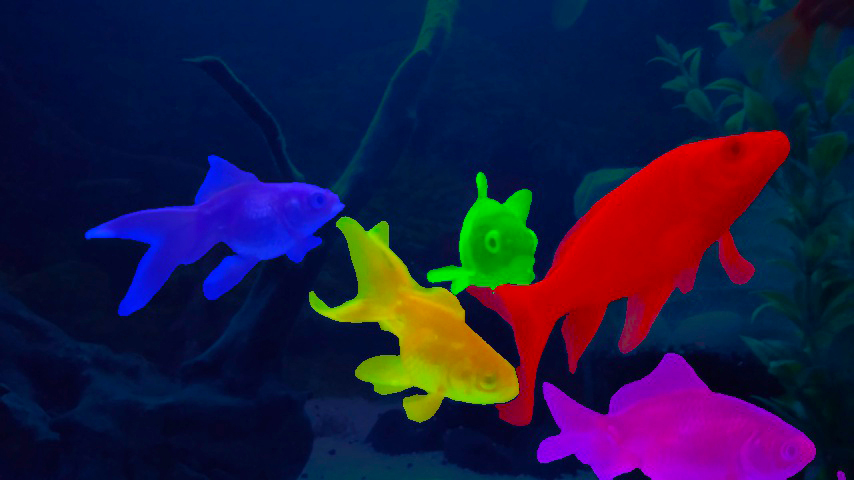}
& \includegraphics[width=0.23\linewidth]{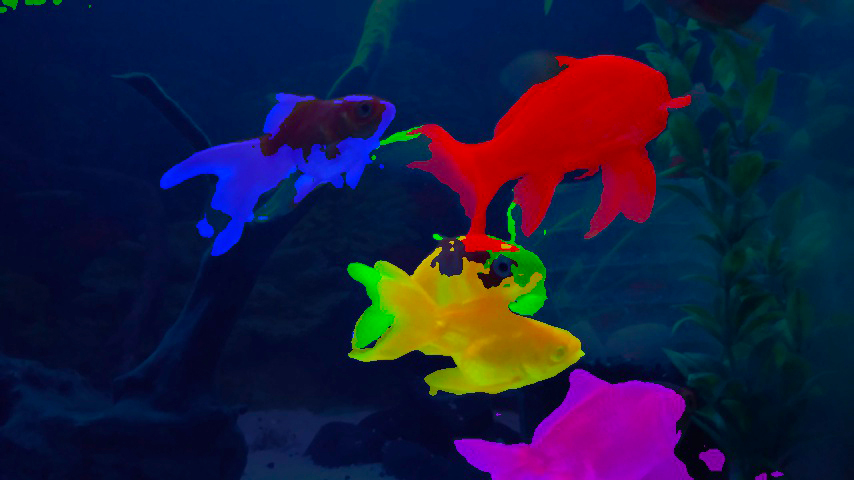}
& \includegraphics[width=0.23\linewidth]{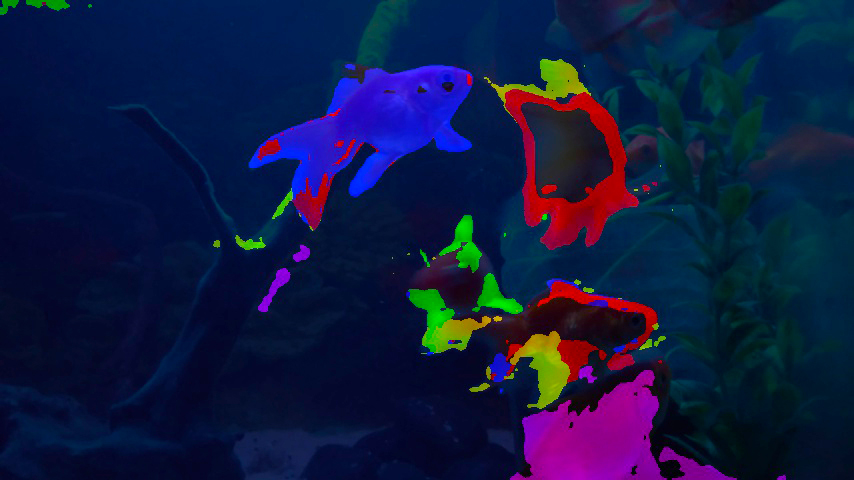}
& \includegraphics[width=0.23\linewidth]{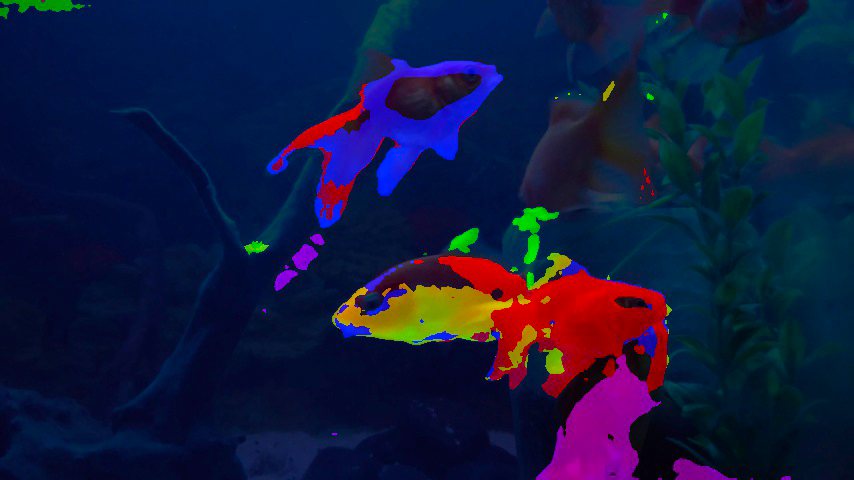}\\
\raisebox{0.1\height}{\rotatebox{90}{\scriptsize OSVOS+}}
& \includegraphics[width=0.23\linewidth]{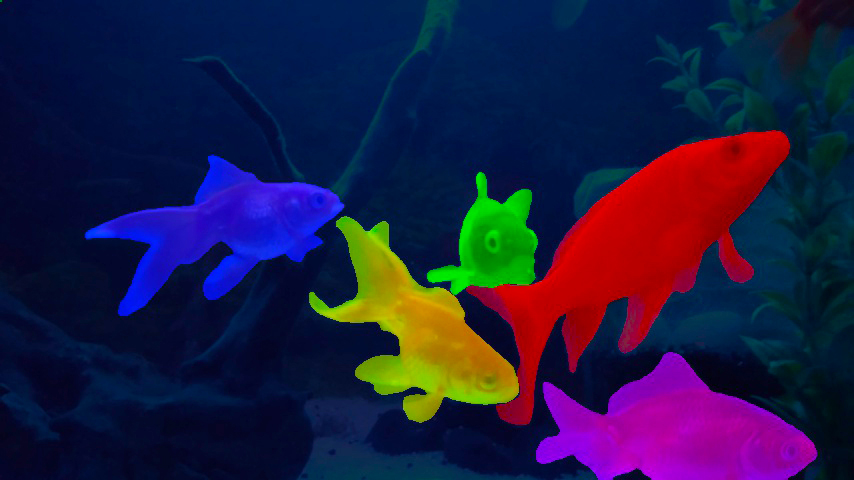}
& \includegraphics[width=0.23\linewidth]{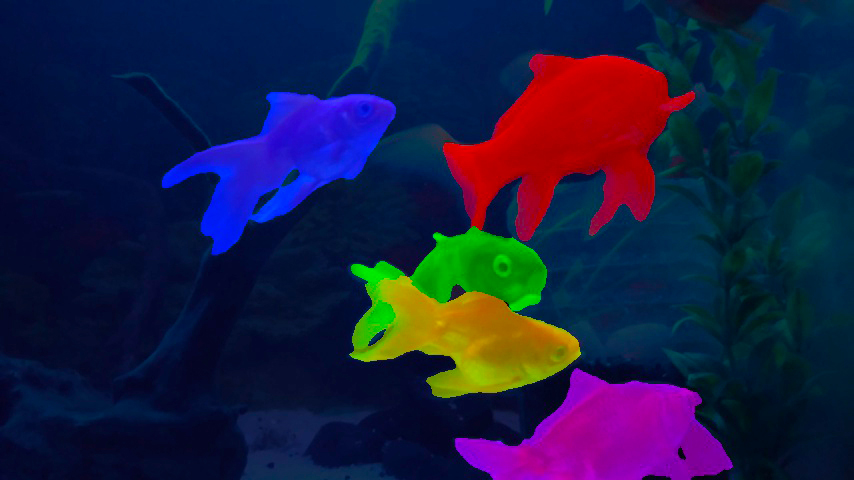}
& \includegraphics[width=0.23\linewidth]{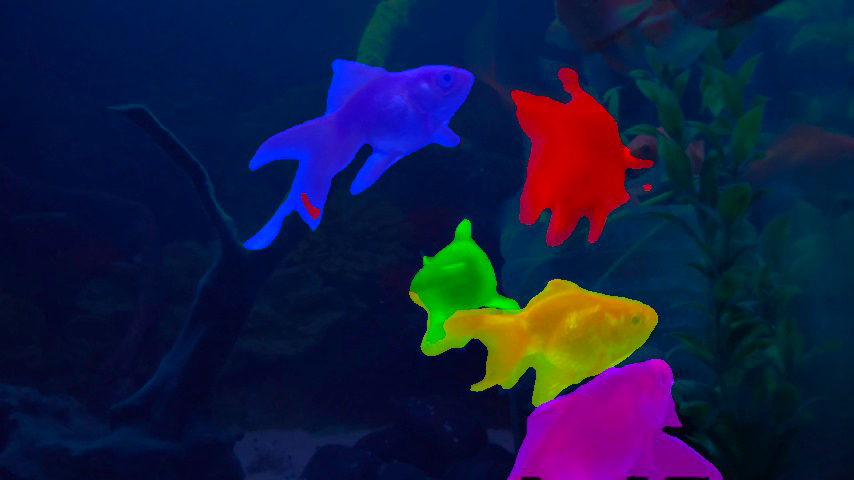}
& \includegraphics[width=0.23\linewidth]{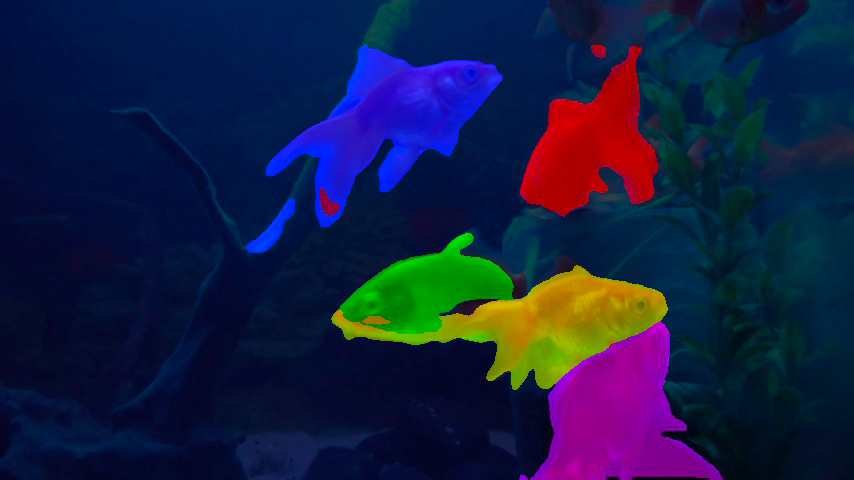}\\
\raisebox{0.3\height}{\rotatebox{90}{\scriptsize RGMP}}
& \includegraphics[width=0.23\linewidth]{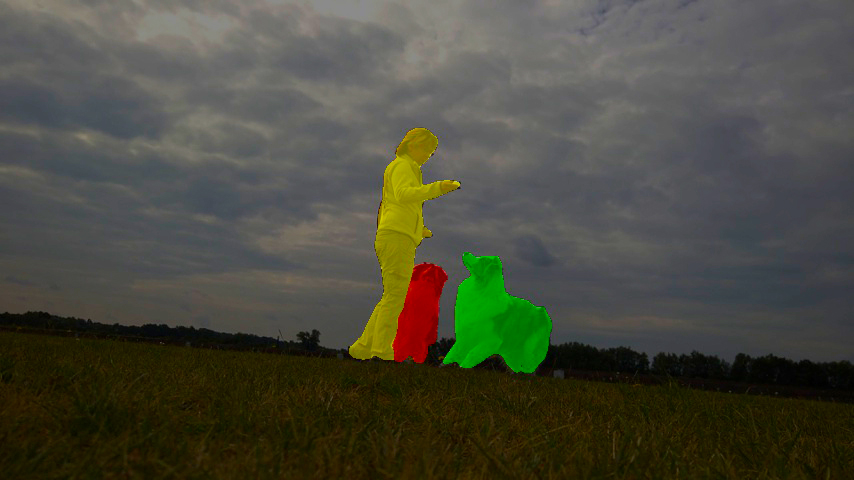}
& \includegraphics[width=0.23\linewidth]{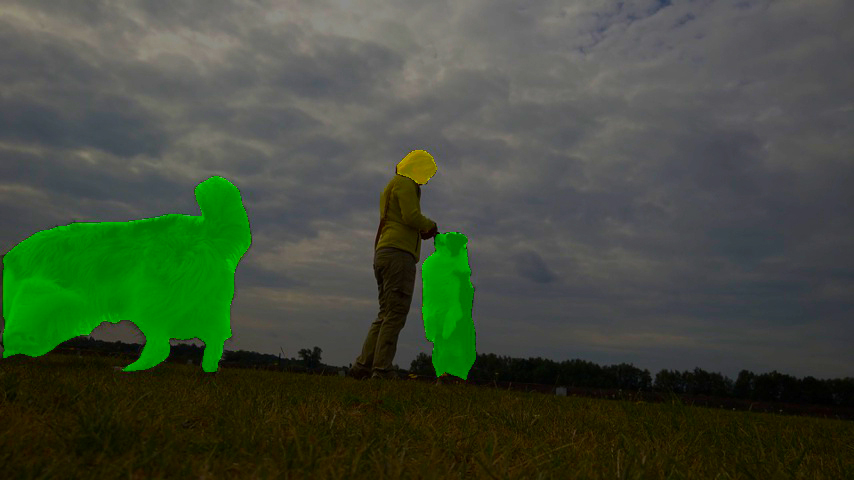}
& \includegraphics[width=0.23\linewidth]{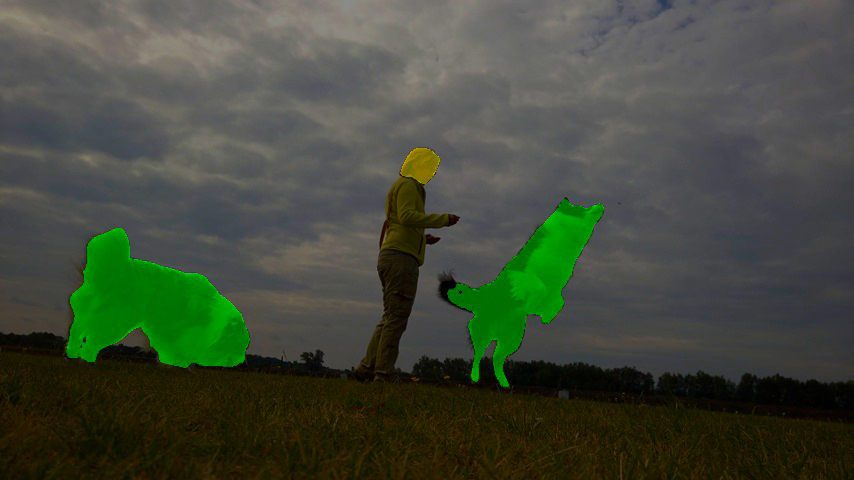}
& \includegraphics[width=0.23\linewidth]{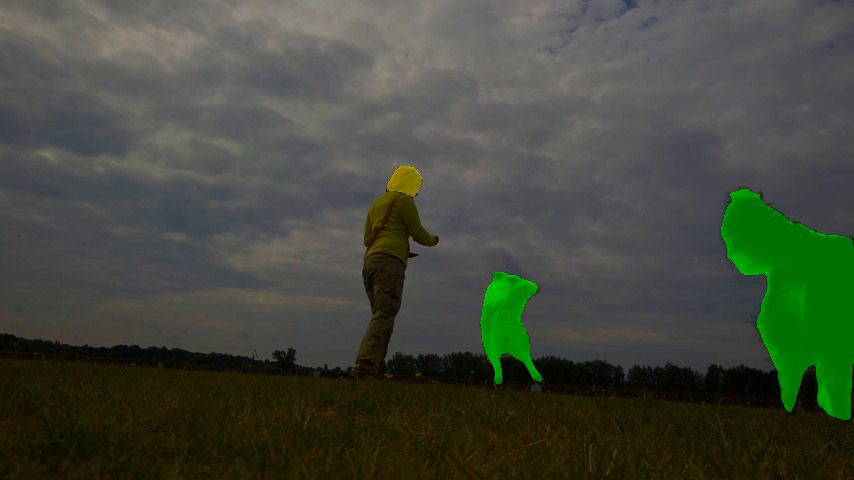}\\
\raisebox{0.2\height}{\rotatebox{90}{\scriptsize RGMP+}}
& \includegraphics[width=0.23\linewidth]{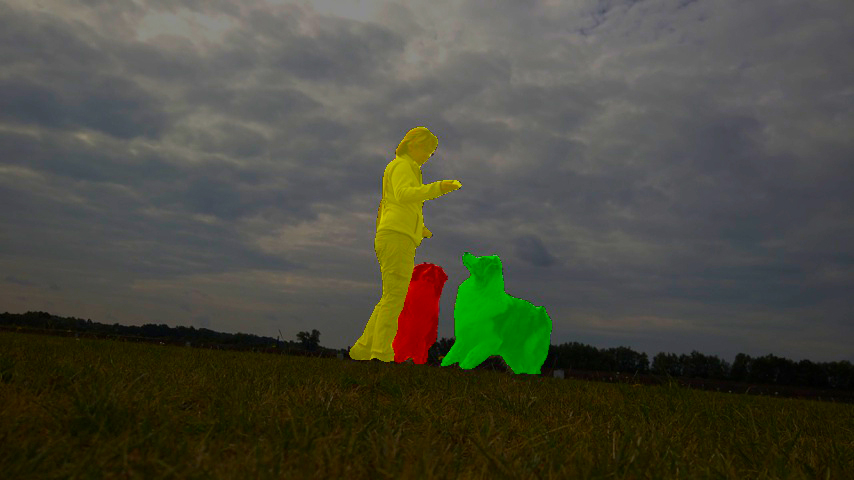}
& \includegraphics[width=0.23\linewidth]{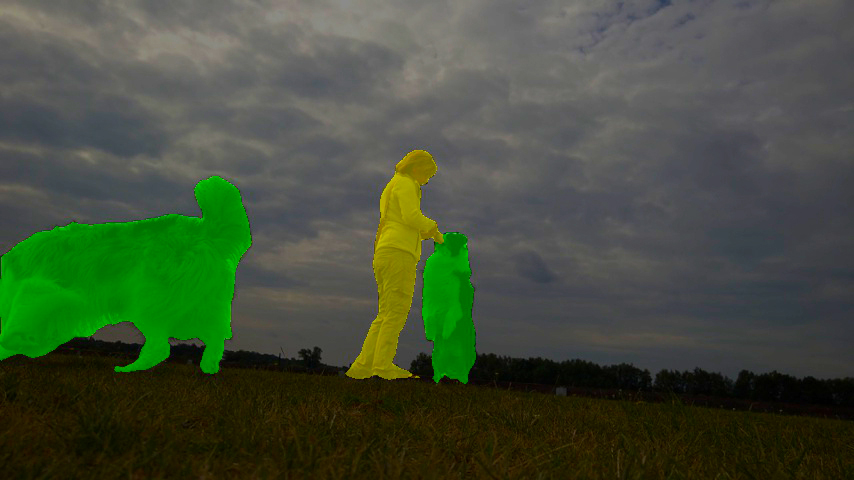}
& \includegraphics[width=0.23\linewidth]{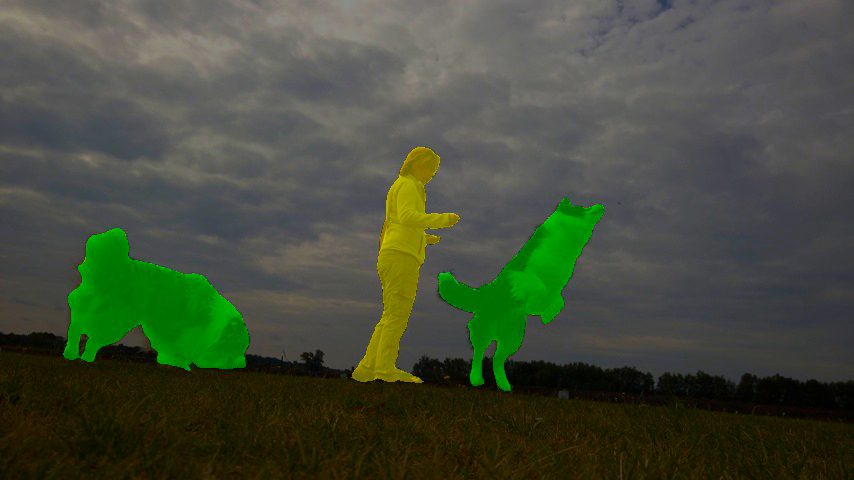}
& \includegraphics[width=0.23\linewidth]{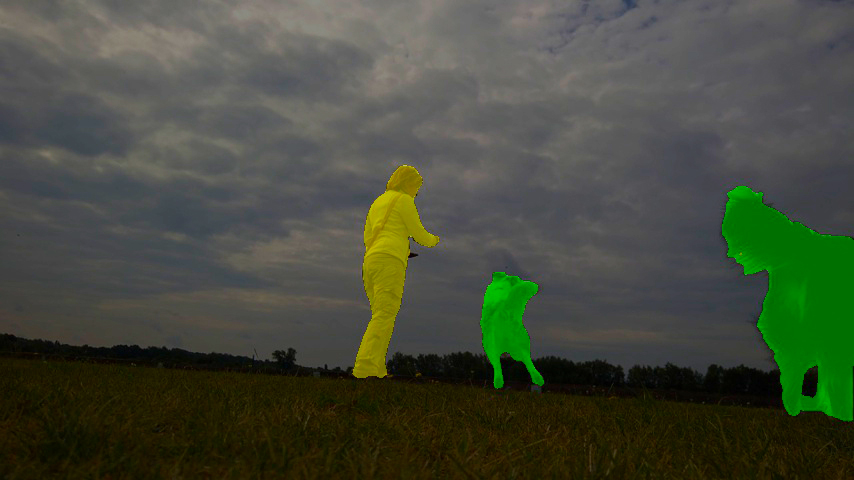}\\

\raisebox{0.5\height}{\rotatebox{90}{\scriptsize STM}}
& \includegraphics[width=0.23\linewidth]{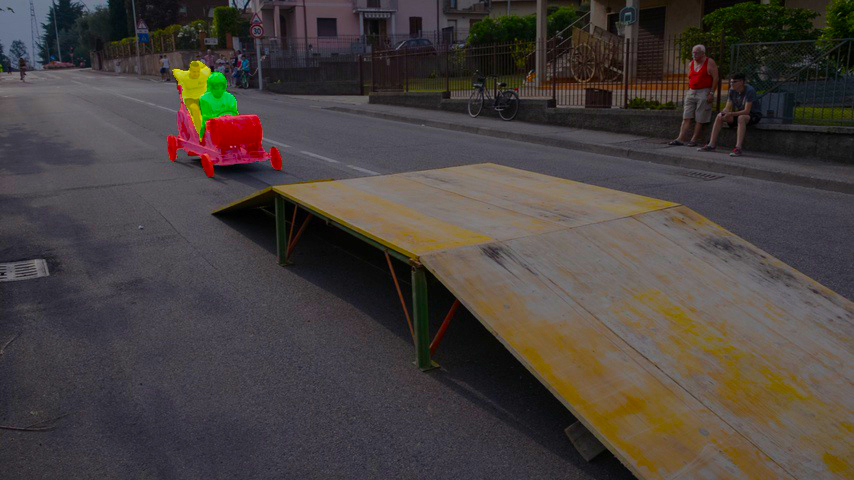}
& \includegraphics[width=0.23\linewidth]{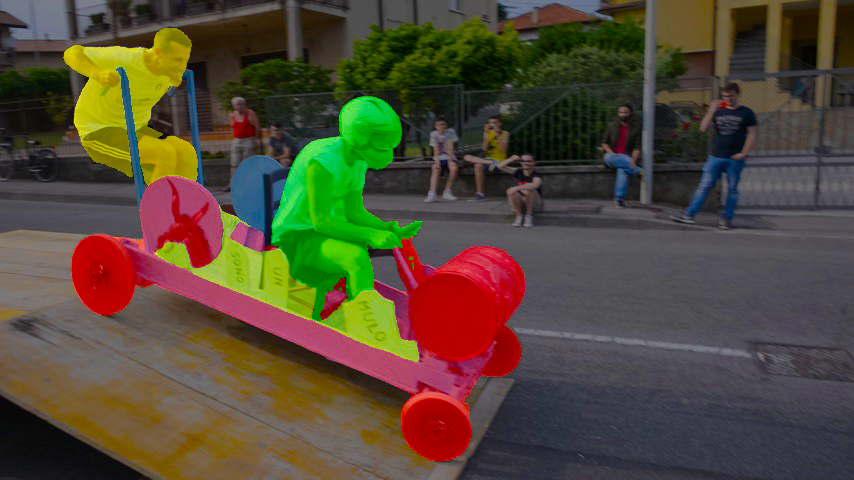}
& \includegraphics[width=0.23\linewidth]{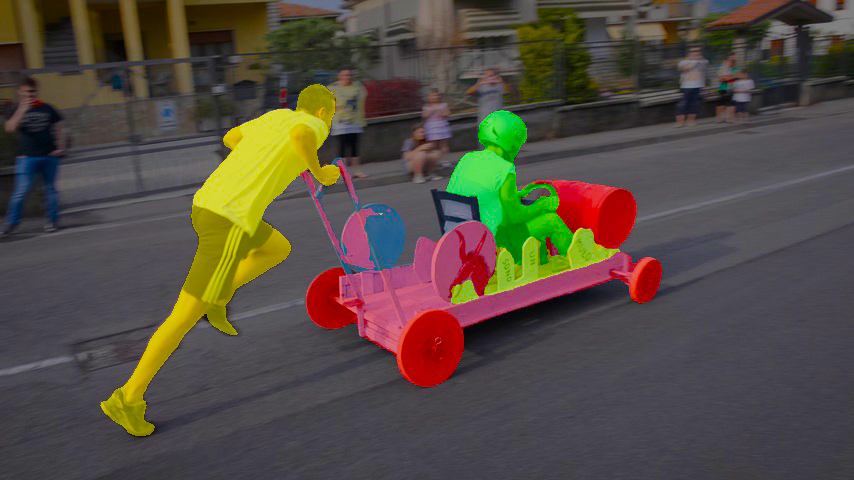}
& \includegraphics[width=0.23\linewidth]{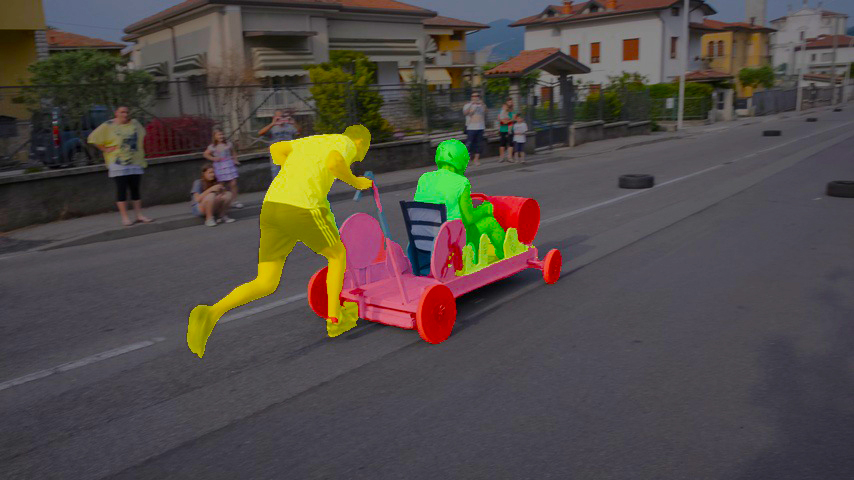}\\
\raisebox{0.3\height}{\rotatebox{90}{\scriptsize STM+}}
& \includegraphics[width=0.23\linewidth]{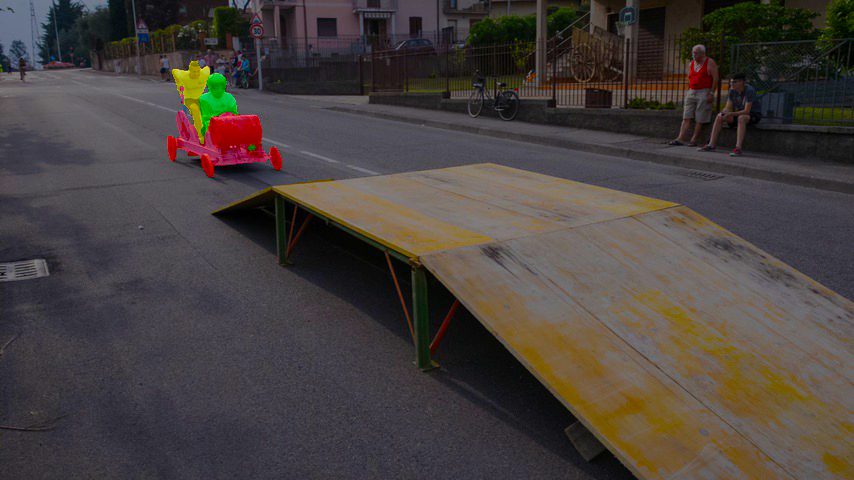}
& \includegraphics[width=0.23\linewidth]{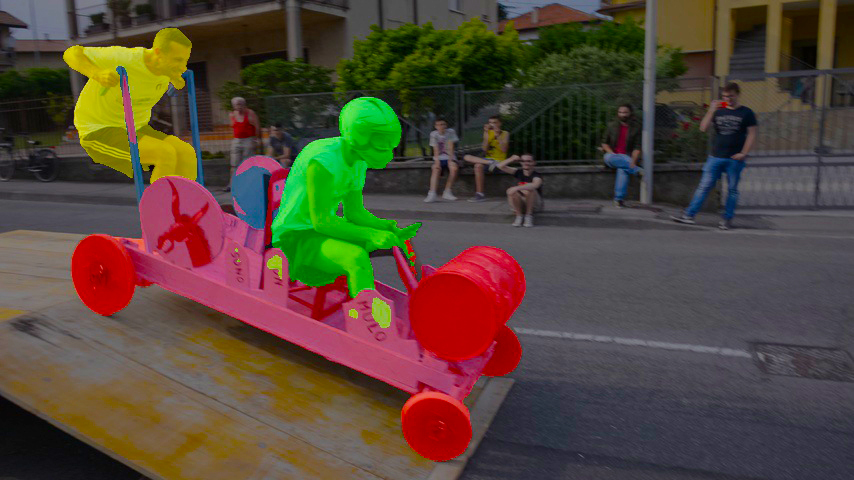}
& \includegraphics[width=0.23\linewidth]{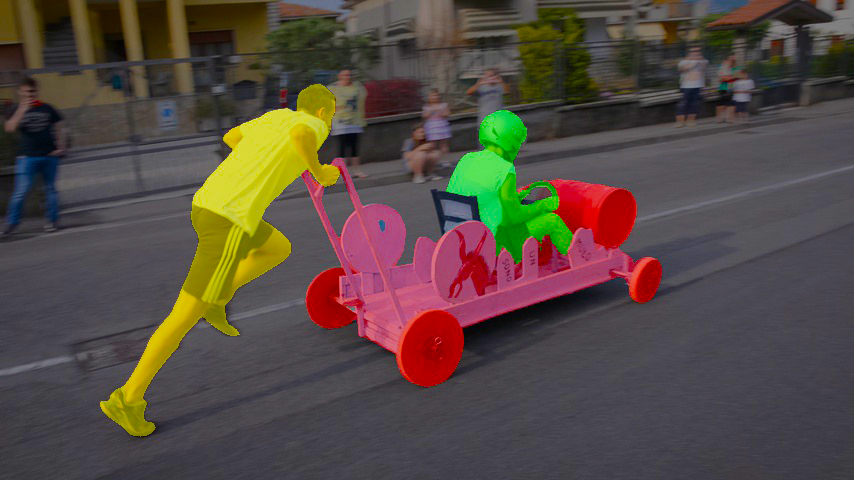}
& \includegraphics[width=0.23\linewidth]{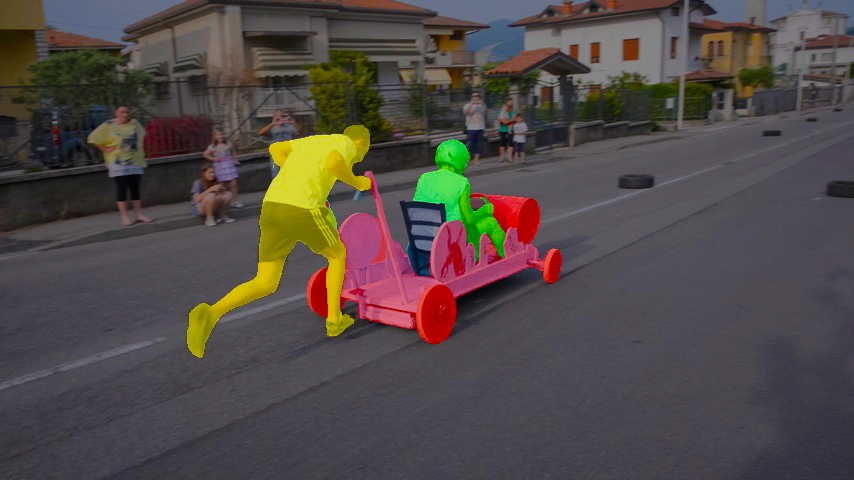}\\
\end{tabular}
\vspace{-9pt}
\caption{The results of OSVOS~\cite{caelles2017one}, RGMP~\cite{oh2018fast}, STM~\cite{oh2019video}, and their improvement by ours (OSVOS+, RGMP+, STM+). }
\label{fig:qualtive}
\vspace{-9pt}
\end{figure}
\subsection{Evaluation and Comparison}
Tab.~\ref{tab:quant} shows the performance gain after applying our approach to three baseline models. By employing the proposed method, OSVOS can outperform all existing online learning-based methods on DAVIS 16 without any post-processing that is widely adopted in existing methods~\cite{maninis2018video,meinhardt2020make}. RGMP and STM with our method also show performance gain on both DAVIS 16 and DAVIS 17 datasets with only 3 extra convolution layers. The experimental results show that the proposed method is generic and widely applicable to various VOS models. 
Fig.~\ref{fig:qualtive} shows the visual examples of the proposed method on three baseline models. The first video contains five similar instances, which are challenging for OSVOS to address. OSVOS with ours (OSVOS+) become more robust towards similar instances under the noisy predictions. In the second video, RGMP with ours (RGMP+) can deal with the partial occlusion of two objects with different semantics. However, our method still fails when the object is fully occluded by the same semantics, which is also a common problem in propagation-based methods. The third video presents a typical failure case of STM, which is caused by scale changes of the target. STM with ours (STM+) can address this issue as well as suppress the error propagation of the video, which improves the long-term robustness of STM. HMMN~\cite{seong2021hierarchical} and STCN~\cite{cheng2021stcn} are recent works that extend STM by improving memory matching. Ours boosts STM to be competitive to HMMN and STCN and can be applied to these methods.
\subsection{Analysis}
In this section, we analyze various components in the proposed method by ablation studies.

\noindent
\textbf{Multi-scale context aggregation.}
Tab.~\ref{tab:ablation} shows the effect of the proposed multi-scale context aggregation module. The overall performance of OSVOS and STM is improved by $1.4\%$ and $2.1\%$, respectively, thanks to the better contour localization of a large input and better perception of the global context of a small input. Noting that RGMP averages the multi-scale prediction of three inputs with different sizes, our module, using only two different-size inputs, can still increase its $\mathcal{J}$ score by $1.3\%$ and $\mathcal{F}$ score by $0.1\%$.

\noindent
\textbf{Scale inconsistency-based variance regularization.} 
In Tab.~\ref{tab:ablation}, we evaluate the effect of our variance regularization on the three backbone models. Our variance regularization shows an average gain of $0.8\%$ for OSVOS. For RGMP and STM, our variance regularization can improve their boundary measurement $\mathcal{F}$ as the scale inconsistency is usually large around the boundary. However, 
considering the RGMP and STM are propagation-based and matching-based methods, the $\mathcal{J}$ score has been slightly degraded as our variance regularization does not enforce the temporal constraint. The proposed self-supervised online adaptation can address this issue by the inter-frame and intra-frame adaptation.

\noindent
\textbf{Inter-frame and intra-frame adaptation.} As shown in Tab.~\ref{tab:addOn}, three baseline models achieve the best performance with both adaptations. Our intra-frame adaptation can encourage the model to learn from the frames without annotations and the inter-frame adaptation can enforce the consistency between the predictions of adjacent frames. These adaptations can reduce the accumulation of the scale inconsistency, which results in improving the temporal stability, $\mathcal{J}$-Decay, and $\mathcal{F}$-Decay. OSVOS, which has poor temporal stability caused by independently processing each frame, shows the most significant improvement.

\begin{table}[t]
\centering
\small

\resizebox{0.78\columnwidth}{!}{
\renewcommand{\arraystretch}{0.9}
\begin{tabular}{c ccc ccc}
\toprule
Backbone & \multicolumn{1}{c}{Ms} & Var & Ada & $\mathcal{J}\uparrow$ & $\mathcal{F}\uparrow$ & $\mathcal{J\&F}\uparrow$ \\
\toprule
& \multicolumn{1}{c}{} & & & 78.0 & 82.6 & 80.3\\
& {\checkmark} & & & 80.0 & 83.4 & 81.7\\
& {\checkmark} & \multicolumn{1}{c}{{\checkmark}} & & 80.4 & 84.6 & 82.5\\
\multirow{-4}{*}{\thead{OSVOS \\ \cite{caelles2017one}}} & {\checkmark} & \multicolumn{1}{c}{{\checkmark}} & \multicolumn{1}{c}{{\checkmark}} & \textbf{86.4} & \textbf{87.9} & \textbf{87.2} \\
\midrule
& \multicolumn{1}{c}{} & & & 81.5 & 82.0 & 81.8\\
& {\checkmark} & & & 82.8 & 82.1 & 82.5 \\
& {\checkmark} & \multicolumn{1}{c}{{\checkmark}} & & 82.4 & \textbf{82.5} & 82.5          \\
\multirow{-4}{*}{\thead{RGMP \\ \cite{oh2018fast}}} & {\checkmark} & \multicolumn{1}{c}{{\checkmark}} & \multicolumn{1}{c}{{\checkmark}} & \textbf{83.1} & 82.4 & \textbf{82.8} \\
\midrule
& \multicolumn{1}{c}{} & & & 88.7 & 90.1 & 89.4\\
& {\checkmark} & & & 91.0 & 91.9 & 91.4\\
& {\checkmark} & \multicolumn{1}{c}{{\checkmark}} & & 90.9 & \textbf{92.0} & 91.4\\
\multirow{-4}{*}{\thead{STM \\ \cite{oh2019video}}}   & {\checkmark} & \multicolumn{1}{c}{{\checkmark}} & \multicolumn{1}{c}{{\checkmark}} & \textbf{91.1} & 91.9 & \textbf{91.5}\\
\bottomrule
\end{tabular}}
\caption{Ablation study of the proposed methods on DAVIS 16. KEY -- Ms: Multi-scale context aggregation, Var: Variance regularization, Ada: Online adaptation.} 
\label{tab:ablation}

\end{table}
\begin{table}[!t]
\centering
\small
\resizebox{0.95\columnwidth}{!}{
\setlength{\tabcolsep}{5pt}
\renewcommand{\arraystretch}{0.9}
\begin{tabular}{cccccccc}
\toprule
Backbone & Intra & Inter & $\mathcal{J}$ $\uparrow$ & $\mathcal{J}$-Decay $\downarrow$& $\mathcal{F}$ $\uparrow$ & $\mathcal{F}$-Decay $\downarrow$& $\mathcal{J\&F}$ $\uparrow$\\ 
\toprule
& & & 62.4 & 32.3 & 69.5 & 31.1 & 66.0 \\
& \checkmark & & 69.9 & 23.9 & 75.8 & 26.8 & 72.8 \\
\multirow{-3}{*}{ \thead{OSVOS \\ \cite{caelles2017one}} } & \checkmark & \checkmark & \textbf{70.8} & \textbf{22.7} & \textbf{76.7} & \textbf{25.6} & \textbf{73.7} \\ \midrule
& & & 64.7 & 20.8 & 69.4 & 23.2 & 67.0 \\
& \checkmark & & 64.9 & 20.0 & 69.6 & 21.9 & 67.3 \\
\multirow{-3}{*}{\thead{RGMP \\ \cite{oh2018fast}}} & {\checkmark} & {\checkmark} & \textbf{65.0} & \textbf{19.6} & \textbf{69.7} & \textbf{21.7} & \textbf{67.4} \\ \midrule
& & & 81.0 & 8.3 & 85.8 & 9.9 & 83.4 \\
& \checkmark & & 81.2 & 8.1 & 85.9 & 9.7 & 83.5 \\
\multirow{-3}{*}{\thead{STM \\ \cite{oh2019video}}} & \checkmark & \checkmark & \textbf{81.3} & \textbf{8.1}  & \textbf{85.9} & \textbf{9.7} & \textbf{83.6} \\ 
\bottomrule
\end{tabular}}
\caption{Ablations studies of our intra-frame adaptation (Intra) and inter-frame adaptation (Inter) on DAVIS 17.}  
\label{tab:addOn}

\end{table}

\begin{table}[!t]
\resizebox{0.95\columnwidth}{!}{
\setlength{\tabcolsep}{5pt}
\renewcommand{\arraystretch}{0.9}
\begin{tabular}{ccccccc} 
\toprule
Backbone & Steps & $\mathcal{J} \uparrow$ & $\mathcal{J}$-Decay $\downarrow$& $\mathcal{F} \uparrow$  & 
$\mathcal{F}$-Decay $\downarrow$& $\mathcal{J\&F}$ $\uparrow$\\  
\toprule  
& 0 & 59.4 & 32.3 & 62.6 & 31.1 & 61.0\\
& 1 & 66.8 & 25.1 & 71.8 & 27.1 & 69.3\\ 
& 10 & \textbf{70.8} & \textbf{22.7} & \textbf{76.7} & \textbf{25.6} & \textbf{73.7}\\ 
\multirow{-4}{*}{\thead{OSVOS \\ \cite{caelles2017one} }}& 20 & 69.4 & 24.0 & 75.3 & 26.5 & 72.3\\ 
\midrule  
& 0 & 64.7 & 20.8 & 69.4 & 23.2 & 67.0\\ 
& 1 & \textbf{65.0} & \textbf{19.6} & \textbf{69.7} & \textbf{21.7} & \textbf{67.4}\\ 
& 3 & 63.7 & 22.8 & 68.9 & 23.8 & 66.3\\ 
\multirow{-4}{*}{\thead{RGMP \\ \cite{oh2018fast}}}& 5 & 63.8 & 21.5 & 68.8 & 23.5 & 66.3\\ %
\midrule  
& 0 & 81.0 & 8.3 & 85.8 & 9.9 & 83.4\\ 
& 1 & 81.2 & 8.1 & 85.9 & \textbf{9.6} & 83.5\\ 
& 3 & \textbf{81.3} & \textbf{8.1} & \textbf{85.9} & 9.7 & \textbf{83.6}\\
\multirow{-4}{*}{\thead{STM \\ \cite{oh2019video}}}& 5 & 81.2 & 8.3 & 85.9 & 10.1 & 83.5\\
\bottomrule 
\end{tabular}}
\caption{Online adaptation with varying steps on DAVIS 17. }  
\label{tab:iterations}
\vspace{-9pt}
\end{table} 

\noindent
\textbf{Iterations for online adaptation.}
We evaluate our self-supervised online adaptation with three baseline models by changing the number of steps for online adaptation. As shown in Tab.~\ref{tab:iterations}, OSVOS can significantly benefit from online adaptation. Since fine-tuning on the first frame and its mask can be limited for online learning-based methods to adapt changes in the video, our online adaptation can provide more training samples to generalize on test sequence. Unlike OSVOS, RGMP is sensitive to the number of iterations as the propagation-based methods are usually hard to train and tend to over-fitting, which leads to performance degradation. Since STM has a memory network to achieve better robustness, STM is not as sensitive to the iterations as RGMP.
\vspace{-7pt}

\section{Conclusion}
We presented a model-agnostic refinement framework for semi-supervised VOS models. The key idea is to improve the pre-trained VOS models by considering the scale-inconsistent predictions from the multi-scale inputs and adapting the models during the test time. Three existing VOS models, with our method, have shown improved segmentation results. Future work includes reducing the computational overhead of our method using meta-learning and alleviating the overfitting issue of our online adaptation.
\vspace{-7pt}

\bibliographystyle{IEEEbib}
\bibliography{icme2022template}

\end{document}